\documentclass{article}

\pdfoutput=1

\usepackage[nonatbib, preprint]{neurips_2024}

\usepackage[utf8]{inputenc} 
\usepackage[T1]{fontenc}    
\usepackage{hyperref}       
\usepackage{url}            
\usepackage{booktabs}       
\usepackage{amsfonts}       
\usepackage{nicefrac}       
\usepackage{microtype}      
\usepackage{xcolor}         
\usepackage{graphicx}
\usepackage[numbers]{natbib}
\usepackage{algorithm}
\usepackage{algpseudocode}
\usepackage{amsmath}

\usepackage{amsthm}
\usepackage{thmtools}

\newtheorem{theorem}{Theorem}
\newtheorem{definition}[theorem]{Definition}

\title{Utilizing Adversarial Examples for Bias Mitigation and Accuracy Enhancement}

\author{%
  Pushkar Shukla * \\
  TTI-Chicago\\
  \texttt{pushkarshukla@ttic.edu} \\
  \And
  Dhruv Srikanth* \\
  Carnegie Melon University\\
  \texttt{dhruvsrikanth5@gmail.com,} \\
  \AND
  Lee Cohen \\
  Stanford University \\
 \texttt{leecohencs@gmail.com} \\
   \And
  Matthew Turk \\
  TTI-Chicago\\
  \texttt{mturk@ttic.edu} \\
}

\begin{document}

\maketitle

\begin{abstract}
\label{sec:abstract}

We propose a novel approach to mitigate biases in computer vision models by utilizing counterfactual generation and fine-tuning. While counterfactuals have been used to analyze and address biases in DNN models, the counterfactuals themselves are often generated from biased generative models, which can introduce additional biases or spurious correlations. To address this issue, we propose using adversarial images, that is images that deceive a deep neural network but not humans, as counterfactuals for fair model training. Our approach leverages a curriculum learning framework combined with a fine-grained adversarial loss to fine-tune the model using adversarial examples. By incorporating adversarial images into the training data, we aim to prevent biases from propagating through the pipeline. We validate our approach through both qualitative and quantitative assessments, demonstrating improved bias mitigation and accuracy compared to existing methods. Qualitatively, our results indicate that post-training, the decisions made by the model are less dependent on the sensitive attribute and our model better disentangles the relationship between sensitive attributes and classification variables.


\end{abstract}

\section{Introduction}
\label{sec:intro}
Computer vision systems trained on large amounts of data have been at the heart of recent technological innovation and growth. Such systems continue to find increasing use-cases in different areas such as healthcare, security, autonomous driving, remote sensing and education. However, in spite of the tremendous promise of large vision models, several studies have demonstrated the presence of unwanted societal biases in these models \cite{bolukbasi2016man,buolamwini2018gender,hendricks2018women}. Therefore, understanding and mitigating these biases is crucial for deploying such applications in real-world scenarios.

Several approaches have been proposed to mitigate \cite{buolamwini2018gender,seyyed2021underdiagnosis,hendricks2018women,meister2023gender,wang2022revise,liu2019fair,joshi2022fair,wang2020towards,wang2023overwriting}, measure \cite{denton2019image,balakrishnan2021towards}, and explain \cite{abid2022meaningfully,feder2021causalm,wu2021polyjuice} biases in computer vision models. Among these approaches, the use of counterfactuals has emerged as a promising and prominent line of research \cite{denton2019image,balakrishnan2021towards,abid2022meaningfully,feder2021causalm,wu2021polyjuice}.  
 A counterfactual for an image, within the context of a model, is a modified version of the original image where certain regions are systematically replaced. This alteration 
 is used to assess 
 the output of the system when presented with the modified image, while ensuring that the modified image remains  similar to the original in terms of its key features and characteristics \cite{vandenhende2022making}. 
\begin{figure}[tb]
  \centering
   \includegraphics[width=\linewidth]{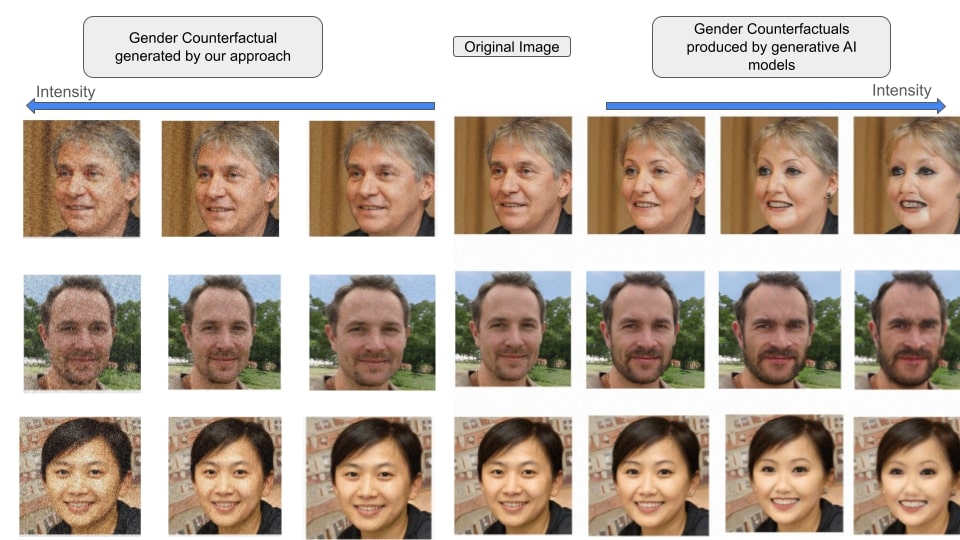}
   \caption{An example of gender counterfactuals produced by StyleGAN2 versus our method, for a smile classification task. StyleGAN2's images (right) correlate femininity with darker lipstick and exaggerated smiling, introducing biases. 
   Our approach (left) generates ASACs that retain the same visual appearance as the original image.}
\label{fig:style_gan2}
\end{figure}

However, current counterfactual generation algorithms used for bias mitigation have several limitations. Firstly, most of these algorithms rely on generative models to produce image counterfactuals with respect to a sensitive attribute (e.g.,\ race, gender). Counterfactuals generated by these image generation mechanisms may inherently contain spurious correlations due to the presence of bias in the image generation model. 

One such example of biased counterfactuals produced by an image generation model, StyleGAN2\cite{karras2020analyzing}, is illustrated for smile classification in Figure \ref{fig:style_gan2}, 
%
where female counterfactuals generated by the model exhibit exaggerated smiles and dark makeup.
Using such counterfactuals to address gender biases in smile prediction may exacerbate the issue they are meant to solve. Additionally, generating counterfactuals based on sensitive attributes like race or gender raises ethical concerns regarding the stereotypes associated with images generated along dimensions of a sensitive attribute. For instance, attributing specific appearances (heavy makeup and increased smile) to female faces may oversimplify their diversity. Therefore, to ensure fairness in computer vision models, we must carefully generate and integrate counterfactuals without introducing new or exacerbating already present biases.

We propose using adversarial images as an alternative to generative models for creating image counterfactuals to assess and mitigate biases in computer vision models. Traditionally designed to deceive models, adversarial images can improve fairness metrics and accuracy. We hypothesize that these specifically designed adversarial images will reduce spurious biases and enhance model robustness and performance. Our method uses existing adversarial techniques to create counterfactuals that challenge vision models based on sensitive attributes. We refer to these examples as ``Attribute-Specific Adversarial Counterfactuals'' or ASACs. Our method ensures that ASACs retain the visual appearance of the original image, effectively addressing ethical concerns regarding the quality of image counterfactuals. By keeping parts of the image unaltered, the likelihood of introducing spurious correlations (propagated by the generative model during the image generation process) into the fairness mitigation pipeline is markedly reduced.


In addition, our work introduces a novel curriculum learning based fine-tuning approach that utilizes these counterfactuals to mitigate biases in vision models. Our approach utilizes ASACs generated at different noise magnitudes and assigns a learning curriculum to these examples based on their ability to deceive the model. We then fine-tune a biased model using the curriculum of ASACs generated. We validate our approach both qualitatively and quantitatively using experiments on various classifiers trained on the CelebA \cite{liu2015faceattributes} and LFW \cite{LFWTech} datasets. Our findings indicate that the proposed training approach not only improves fairness metrics but also maintains or enhances the overall performance of the model. In addition, we show that our method generalizes to biased models of varying scales (6M-24M). Our contributions can be summarized as follows: we introduce a bias-averse method for generating image counterfactuals using adversarial images to create Attribute-Specific Adversarial Counterfactuals (ASACs) for sensitive attributes. We also present a novel curriculum learning-based fine-tuning approach that leverages ASACs to mitigate pre-existing biases in vision models. Additionally, we provide a comprehensive evaluation across various datasets, architectures and metrics, demonstrating that our method enhances fairness metrics without compromising the model's overall performance.

\section{Related Works}
\label{sec:related}

\begin{figure}[tb]
  \centering
   \includegraphics[width=\linewidth]{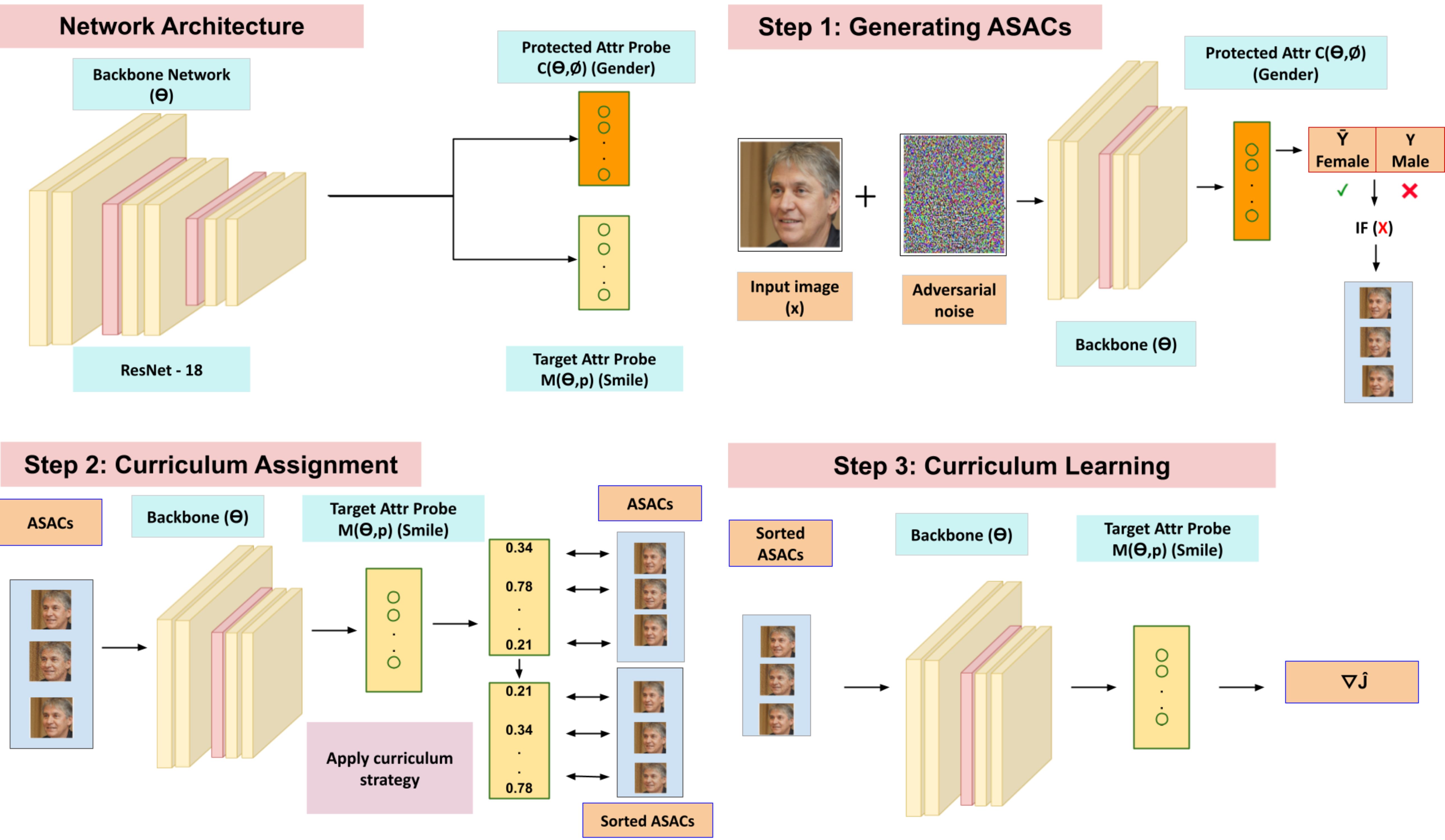}
   \caption{\textbf{Bias Mitigation Strategy}: Our proposed solution for mitigating biases in a model (e.g., smile classifier) $M(\theta,\rho)$ involves training sensitive attribute classifier  $C(\theta,\phi)$ (shown in the network architecture). We then follow a three-stage pipeline. (1) We generate ASACs that are capable of deceiving $C(\theta,\phi)$. (2) We define a curriculum assignment strategy that organizes these ASACs on based on the degree to which they deceive the original model $M(\theta,\rho)$. (3) We fine-tune the original model $M(\theta,\rho)$ using the organized ASACs and the proposed loss function (see Equation \ref{eq:adv_loss}).
}
\label{fig:three_step_pipeline}
\end{figure}
\paragraph{\textbf{Adversarial examples}}

Adversarial examples have emerged as a notable challenge in the realm of computer vision\cite{maudslay2019s,wang2021adversarial,szegedy2013intriguing}. These examples are crafted with the specific goal of deceiving machine learning models by making subtle alterations to input data, resulting in model misclassification. Adversarial attacks are specialized algorithms designed to generate such examples and have been the subject of extensive research. These attacks are categorized into black-box \cite{rahmati2020geoda,jiang2019black,narodytska2017simple} and white-box attacks \cite{szegedy2013intriguing,maudslay2019s}, depending on the attacker's access to model parameters. Our approach uses two white-box methods, FGSM \cite{goodfellow2014explaining} and PGD\cite{madry2017towards}.

\paragraph{\textbf{Bias mitigation in computer vision}}
Bias mitigation strategies in computer vision models can be categorized into three main approaches depending on when they are deployed into the machine learning pipeline: pre-processing, in-processing, and post-processing approaches. Pre-processing methods \cite{quadrianto2019discovering,ramaswamy2021fair,zhang2020towards} focus on addressing biases before training the models. This is achieved by either adjusting the data distribution or strategically augmenting existing data. In-processing techniques\cite{roh2020fr,shi2020towards,zeng2023deep} try to improve the model during the training phase either by proposing an alternate training strategy or by changing the model structure. Post-processing strategies \cite{wang2022fairness,lohia2019bias} adjust fairness criteria after the model is trained. Our approach falls within the post-processing category as it is a method to fine-tune an already existing model concerning a protected attribute.

\paragraph{\textbf{Counterfactuals and their role in machine learning}}
 The notion of using counterfactuals in machine learning models is closely related to the definition of counterfactual fairness proposed by Kusner et al. \cite{kusner2017counterfactual}. Counterfactuals have emerged as a versatile tool in machine learning \cite{kasirzadeh2021use,mothilal2020explaining,sokol2019counterfactual}, natural language processing  \cite{chen2023disco,kaushik2019learning}, and computer vision \cite{abid2022meaningfully,feder2021causalm,wu2021polyjuice,abbasnejad2020counterfactual,denton2019image,balakrishnan2021towards,dash2022evaluating,cheong2022counterfactual,wang2020towards,zhang2020towards}. Counterfactual images have been used to explain  \cite{abid2022meaningfully,feder2021causalm,wu2021polyjuice}, analyze \cite{abbasnejad2020counterfactual,denton2019image,balakrishnan2021towards,chinchure2023tibet} and improve fairness \cite{dash2022evaluating,cheong2022counterfactual} in computer vision models.

\paragraph{\textbf{The Relationship Between Counterfactual and Adversarial Examples}}
While the connection between counterfactuals and adversarial images has been extensively explored in theoretical machine learning\cite{pawelczyk2022exploring,ustun2019actionable,van2021interpretable, karimi2020model,beutel2017data}, limited attention has been given to this relationship in computer vision\cite{zhang2020towards,qiu2020semanticadv,lim2023biasadv,wang2020score}. Among these approaches, work by Wang et al. \cite{wang2020towards} and Zhang et al.\cite{zhang2020towards} is particularly noteworthy because of its proximity to our approach. Wang et al. proposed a post-processing approach to mitigate biases using adversarial examples, uniquely incorporating a GAN-based loss function. Conversely, Zhang et al. employ an architecture similar to ours, generating adversarial perturbations to counteract biases. Our research however, is distinct in proposing a novel training setup that introduces a curriculum learning strategy along with a loss objective for fine-grained control. Unlike previous methods that often rely on a single adversarial example, our approach systematically assesses the impact of each adversarial example on the original classifier, providing a more directed training regime.

\paragraph{\textbf{Curriculum Learning}} Curriculum learning is an effective training paradigm that gradually exposes models to progressively more complex examples, aiding in better generalization \cite{bengio2009curriculum,jiang2015self,matiisen2019teacher,kong2021adaptive,graves2017automated}. This approach strategically organizes training data, facilitating smoother convergence and improved performance. Several approaches have since been proposed to organize data that include sorting the data \cite{bengio2009curriculum}, adaptive organization \cite{matiisen2019teacher}, self-paced curriculum assignment \cite{matiisen2019teacher}, and teacher-based curriculum learning \cite{matiisen2019teacher}.


\section{Method}
\label{sec:method}
\paragraph{Notations}
\label{sec:notation}
We define the data distribution as $D$, from which image-label pairs $(x,y)$ are sampled \textit{i.i.d.}, both for train and test data. Protected attribute pairs are denoted as $(x_a, y_a)$ and are sampled from the same data distribution as $(x,y)$ with adversarial perturbations relative to an attribute-specific image $x_a$ represented by $\delta_{x}$. As an example consider a smile classifier with gender as the protected attribute. Here $(x,y)$ is the input and true label of the smile classifier whereas $x_a$ denotes the image when the input image is $x$  is perturbed. We define the target attribute classifier as $M(\theta, \rho)$, using a backbone network $\theta$ and a linear probe $\rho$, while the protected attribute classifier as $C(\theta, \phi)$, sharing the backbone $\theta$ but with a different linear probe $\phi$. Predictions from these classifiers are $\hat{Y} = M(\theta, \rho, x)$ for the target and $\bar{Y} = C(\theta, \phi, x_a)$ for the protected attribute. 


Our method for creating attribute-specific adversarial counterfactuals (ASACs) and employing them to fine-tune a biased model can be broken down into three stages. First, we generate ASACs for a set of input images $X= \{x^1, x^2, \ldots,x^n\}$, aiming to mislead the classification model $M(\theta, \rho)$ regarding a protected attribute $a$ (detailed in Section \ref{sub:gen_ASAC}). For example, when training a smile classifier, we produce adversarial images that prompt misclassification based on the protected attribute of gender. The next step (Section \ref{sub:curriculum}) evaluates the effectiveness of the generated ASACs. Continuing with our example, we measure the capacity of the ASACs to incorrectly influence the model's smile classification, despite the ASACs being generated to confuse the model about gender. We assign a training curriculum based on the ASAC's success in deceiving the model. This procedure can be seen in Algorithm \ref{alg:asacs_curriculum}. Finally, the concluding stage utilizes the curriculum and ASACs produced, and introduces an adversarial loss (Section \ref{sub:fine_control_adv_loss}) to fine-tune the original model $M(\theta, \rho)$. This aims to reduce biases within the model and enhance its discriminative performance. The training setup and hyperparameters are detailed in Section \ref{sup:train_setup}.

\subsection{Generating attribute-specific adversarial examples}
\label{sub:gen_ASAC}
This section details our approach to constructing ASACs - adversarial images designed to mislead a model concerning a specific sensitive attribute, denoted as $a$. We start by employing a target classifier $M(\theta, \rho)$ and develop an attribute classifier, $C(\theta, \phi)$, to predict the protected attribute $a$ within the dataset $D$. Both models share a similar structure, with only the last layer differing, as depicted in Figure \ref{fig:three_step_pipeline}, thereby utilizing nearly identical representations for classification. They are trained on identical data distributions $D$. To construct each ASAC $x_a$, we augment the original image $x$ with noise $\delta_{x_a}$ using common methods to generate adversarial images as shown below.

\begin{equation}
\label{eq:adv_examples}
    x_a= x + \delta_{x_a}
\end{equation}
\begin{algorithm}[H]
    \caption{Construct Minibatch for Training based on Curriculum}
    \label{alg:asacs_curriculum}
    \begin{algorithmic}[1]
        \Function{ConstructCurriculum}{$M(\theta, \rho)$, $\{x_a(1), \dots , x_a(k)\}$, $\{\epsilon(1), \dots , \epsilon(l)\}$, $f_{attack}$, \text{order}}
            \Require $M(\theta, \rho)$: Target attribute classifier
            \Require $\{x_a(1), \dots , x_a(k)\}$: Attribute-specific minibatch
            \Require $\{\epsilon_1, \dots , \epsilon_l\}$: Curriculum of noise magnitudes
            \Require $f_{attack}$: Any scalable adversarial attack
            \Require \text{order}: Order to sort ASACs in
            \State $\text{minibatch} \gets \{\}$
            \For{$i \gets 1$ \textbf{to} $k$} 
                \For{$j \gets 1$ \textbf{to} $l$}
                    \State $\mathbf{x'_a} \gets f_{attack}(x_a(i), \epsilon_j)$  \Comment{Construct ASAC with Eq. \ref{eq:adv_examples}}
                    \State $\text{score} \gets DS(\mathbf{x'_a}; M(\theta, \rho))$ \Comment{Compute Difficulty Score with Eq. \ref{eq:ds_2}}
                    \State $\text{minibatch}.\text{append}(\mathbf{x'_a}, \text{score}))$
                \EndFor
            \EndFor
            \State $\text{minibatch}.\text{sort}(key=\text{lambda } x: x[1])$ \Comment{Sort minibatch by difficulty score}
            \State \textbf{return} $\text{minibatch}$
        \EndFunction
    \end{algorithmic}
\end{algorithm}

In our work, we experiment with two commonly used adversarial methods, Fast Gradient Sign Method (FGSM) \cite{goodfellow2014explaining} and Projected Gradient Descent (PGD) \cite{madry2017towards}, to generate images capable of deceiving the attribute classifier. FGSM is a single-step attack approach that perturbs input images based on the gradient sign of the loss function, as shown in Equation \ref{eq:fgsm}, with $\epsilon\in[0,1]$ denoting a scaling factor which we refer to as the noise magnitude. PGD iteratively perturbs input data similar to FGSM to maximize the loss function, aiming to find the smallest perturbation that causes misclassification.

\begin{equation}
\label{eq:fgsm}
\delta_x = \varepsilon \times \text{sign} \left( \nabla_x J(\theta, \phi, x_a, y) \right)
\end{equation}

 

\subsection{Curriculum Learning} 
\label{sub:curriculum}
We propose a curriculum learning based strategy for training our model. Our approach comprises two main stages. First, we evaluate the influence of each ASAC on the classification model $M(\theta,\phi)$ by assessing its difficulty score. Subsequently, the model implements a curriculum for training, guided by the difficulty scores of all examples. 

\subsubsection{Computing Difficulty Scores}
\label{sub:difficulty}
The difficulty score is crucial in curriculum learning, as it measures the utility of each example and determines the order in which ASACs are presented to the model during adversarial training \cite{kong2021adaptive}. However, obtaining the optimal difficulty score a priori is not feasible in our setting. To address this, we define the difficulty score for any attribute-specific image $x_a$ as the complement of the softmax value that the model $M$ returns for $x_a$.

\begin{equation}
\label{eq:ds_1}
    DS(x_a) = 1 - \text{Softmax}(x_a; M_{\theta,\rho}) = 1 - \frac{\exp(M_{\theta,\rho}(x_a))}{\sum_{i=1}^{N}\exp(M_{\theta,\rho}(x_a(i)))}
\end{equation}

\noindent where $N$ is the number of classes for the target outcome. Note that here we use ASACs for adversarial curriculum learning, therefore, we  rewrite Equation (\ref{eq:ds_1}) as shown below:
\begin{equation}
\label{eq:ds_2}
    DS(x'_a) = 1 - \text{Softmax}(x'_a; M_{\theta,\rho})
\end{equation}

A subtle but key point to note here is that the model used to determine the difficulty score and ultimately the curriculum is the target classifier $M(\theta, \rho)$, however, the examples being scored are the ASACs generated, i.e.,\ $x'_a$ which are generated as a function of the protected attributed classifier $C(\theta, \phi)$. 

\subsubsection{Curriculum Assignment}
\label{sub:cl_fair_adv_learning}

For each minibatch of $k$ attribute-specific images $\{x_a(1), \dots , x_a(k)\}$, we compute the Cartesian product of ASACs with the set of $l$ noise magnitudes we use in the curriculum (including $\epsilon = 0$). For each of the samples in the resulting minibatch of size $k \times l$, we compute a difficulty score, $DS(x'_a)$, detailed in Equation \ref{eq:ds_2}, that scores an ASAC based on the difficulty of the example for the target classifier to classify correctly. We then sort the minibatch based on the difficulty of each sample. See Table \ref{tab:cl_order} for an ablation on the effect of a randomized, increasing, and decreasing order (\textit{w.r.t.} difficulty score) for samples in the curriculum. This procedure is illustrated in detail in Algorithm \ref{alg:asacs_curriculum} and enables what defines each in-batch curriculum for corresponding minibatches in the fine-tuning process of the target classifier $M(\theta, \rho)$.

\subsection {Fine-Grained Control using Adversarial Loss}
\label{sub:fine_control_adv_loss}

It is important to note that in all curriculums experimented with, we include the original minibatch, i.e.,\ $\epsilon = 0$. When performing any kind of adversarial training, there exists the possibility of overfitting, especially when fine-tuning a model over the same or similar data it was originally trained on. Though overfitting may lead to improved fairness metrics, it results in a lower degree of generalization of the model. While fine-tuning the target classifier $M(\theta, \rho)$, we use \textit{Cross Entropy} (CE)  as our loss function ($J$);  however, any fully differentiable loss function may be used. We extend our approach to address the issue of overfitting and the ability to have fine-grained control over the trade-off between accuracy and fairness by introducing an adversarial regularization term, $\alpha$. We construct our loss function $\hat{J}$ to contain two terms, both governed by a convex combination of $\alpha$ as shown in the equation below:
\begin{equation}
    \label{eq:adv_loss}
\hspace{-0.5 cm}
\begin{aligned}
    \hat{J}_\alpha(\theta, \rho, x_a;x'_a, y) = \alpha J(\theta, \rho, x_a, y) 
    + (1 - \alpha)J(\theta, \rho, x'_a,y) 
\end{aligned}
\end{equation}

\section{Results}
\label{sec:results}

In this section, we present and discuss various experiments to evaluate and interpret our proposed approach. We evaluate our approach through quantitative performance metrics and fairness evaluations, followed by qualitative analyses using visualization methods for model understanding such as Integrated Gradients (IG) \cite{sundararajan2017axiomatic}. In addition, we conduct ablations on different curriculum policies, noise magnitudes ($\epsilon$) and network architectures. We use two popular datasets, CelebA \cite{liu2015faceattributes} and Labeled Faces in the Wild (LFW) \cite{LFWTech}. To evaluate the fairness of our model, we employ three key metrics: the Difference in Demographic Parity (DDP), the Difference in Equalized Odds (DEO), and the Difference in Equalized Opportunity (DEOp) \cite{hardt2016equality}. Alongside these, we also measure the model's accuracy (ACC). Lower values for fairness metrics indicate a less biased model whereas a high value in accuracy indicates a more performant model. For more details about the metrics and their definition please refer to the supplementary material (Section \ref{sup:Deifintion}
). 


\subsection{Quantitative Results}
\label{sub:quant_results}

\begin{table}[t]
\caption{A comparison of our approach with other approaches \cite{zhang2020towards,ramaswamy2021fair} on the CelebA dataset. We also test on a variety of backbone architectures.}
\label{tab:celeba_results}
\vspace*{-0.15in}
\begin{center}
\begin{small}
\begin{tabular}{llrrrr}
\toprule
Label & Method  & DEO & DEOp &DDP& ACC\\
\midrule
\midrule
Smile & Base Classifier  &0.088 &0.066&	0.15&	84.29  \\
&Adversarial Training \cite{zhang2020towards}	&0.077&	0.051&	0.17	& 91.74 \\
&Counterfactual GAN based training \cite{ramaswamy2021fair}  & 0.065 &0.050  &0.17  &  86.50 \\
&ASAC (with FGSM)&\textbf{0.050} &\textbf{0.043}	&0.15	&\textbf{91.91}\\
&ASAC (with PGD )&0.058 &0.045 &\textbf{0.14}	&91.20\\

\midrule
\midrule
Big Nose & Base Classifier  & \textbf{0.332} & \textbf{0.257} & \textbf{0.28} & 80.77  \\
&Adversarial Training \cite{zhang2020towards} & 0.396 & 0.311 & 0.35 & 81.53 \\
&Counterfactual GAN based training \cite{ramaswamy2021fair}  &0.392  &0.301  &0.36 &  78.77 \\
&ASAC (with FGSM) & 0.354 & 0.267 & 0.30 & \textbf{82.05}\\
&ASAC (with PGD) & 0.374 & 0.281 & 0.31 & 81.03\\

\midrule
\midrule
Wavy Hair & Base Classifier & 0.347 & 0.237 & 0.35 & 81.15\\
&Adversarial Training \cite{zhang2020towards}    & 0.359 & 0.264 & 0.40 & \textbf{82.07}\\
&Counterfactual GAN based training \cite{ramaswamy2021fair}  &0.344  &0.230  &0.34 &  79.21 \\
&ASAC (with FGSM) &\textbf{0.34} & \textbf{0.29} & 0.342 & 82.01\\
&ASAC (with PGD) &0.383 & 0.243 &\textbf{ 0.34} & 81.99\\

\bottomrule
\end{tabular}
\end{small}
\end{center}
\vskip -0.1in
\end{table}

\begin{table}[t]
\caption{Performance across accuracy and fairness metrics on the LFW dataset. }
\label{tab:lfw_results}
\vspace*{-0.15in}
\begin{center}
\begin{small}
\begin{tabular}{llrrrr}
\toprule
Label & Method & \hspace{0.5mm} DEO & \hspace{0.5mm} DEOp & \hspace{0.5mm} DDP & \hspace{0.5mm} ACC \\
\midrule
\midrule
Smile & Base Classifier  &0.074 &0.071&	0.19&	71.29  \\
& Adversarial Training \cite{zhang2020towards}	&0.065&	0.059&	0.18	& 84.74 \\
&ASAC (with FGSM)&\textbf{0.054}	&\textbf{0.050}	&\textbf{0.16}	&\textbf{89.19}\\
& \text{ASAC (with PGD)}&0.063 &0.060&	0.16&	88.41    \\
\midrule
\midrule
Bags Under Eyes & Base Classifier  &0.074 &0.073&	0.16&	69.29  \\
&Adversarial Training \cite{zhang2020towards}	&0.064&	0.061&	\textbf{0.14}	& 74.27 \\
&ASAC (with FGSM)&\textbf{0.061}	&\textbf{0.060}	&\textbf{0.14}	&\textbf{87.29}\\
& \text{ASAC (with PGD)} &0.066 &0.062&	0.17&	86.42\\
\midrule
\midrule
Wavy Hair &Base Classifier  &0.082 &0.072&	0.18&	63.29  \\
& Adversarial Training \cite{zhang2020towards}	&\textbf{0.065}&	0.064&	0.19	& 73.27 \\
&ASAC (with FGSM)&0.066	&\textbf{0.062}	&0.19	&88.29\\
&\text{ASAC (with PGD)} &\textbf{0.065} &0.067&	\textbf{0.16}&	\textbf{88.70} \\
\bottomrule
\end{tabular}
\end{small}
\end{center}
\vskip -0.1in
\end{table}

\begin{table}[t]
\caption{Comparison of accuracy and fairness metrics between base classifier and fine-tuning using the proposed approach across different backbone architectures.}
\label{tab:backbone_results}
\vspace*{-0.15in}
\begin{center}
\begin{small}
\begin{tabular}{llrrrr}
\toprule
Model & Backbone Network & DEO & DEOp &DDP& ACC\\
\midrule
\midrule

Base Classifier & ResNet-18 &0.088 &0.066 &0.150 &84.29 \\
Base Classifier & ResNet-50 &0.077 &0.064 &0.148 &86.60\\
Base Classifier & DenseNet-121 &0.090  &0.075  &0.146 &80.36 \\
Base Classifier & DenseNet-169 &0.092 &0.075 &0.139 &80.95\\
ASAC & ResNet-18 &0.050 &0.043 &0.150 &\textbf{91.91}\\
ASAC & ResNet-50 &\textbf{0.047} &\textbf{0.042} &\textbf{0.137} &90.78\\
ASAC & DenseNet-121 &0.052 &0.046 &0.153 &91.32\\
ASAC & DenseNet-169 &0.070 &0.051 &0.158 &91.08\\

\bottomrule
\end{tabular}
\end{small}
\end{center}
\vskip -0.1in
\end{table}

\paragraph{\textbf{Comparing the proposed method across different target attributes on different datasets}} As shown in Table \ref{tab:celeba_results}, we quantitatively evaluate the performance and fairness of our approach on the CelebA\cite{liu2015faceattributes} dataset, focusing on target attributes such as Smile, Big Nose, and Wavy Hair, with gender as the protected attribute. Using ResNet-18\cite{he2016deep} as the backbone, we compare models trained with and without our proposed approach and evaluate against the original baseline model and two different methods: an adversarial training approach \cite{zhang2020towards} and a debiasing approach by Ramaswamy et al.\cite{ramaswamy2021fair} that generates counterfactual images using GANs. Our approach, utilizing FGSM \cite{goodfellow2014explaining} and PGD \cite{madry2017towards} adversarial methods (with $\epsilon=\{0,0.01,0.001\}$) and $\alpha=0.5$, outperforms these previous methods in terms of both accuracy and fairness metrics. While the GAN-based counterfactual approach \cite{ramaswamy2021fair} appears to reduce the overall accuracy of the system post training, our method not only improves accuracy for all three target attributes, but also improves fairness metrics (for all target attributes except big nose). We conduct the same analysis on the LFW dataset \cite{LFWTech}, targeting the attributes Smiling, Bags Under Eyes, and Wavy Hair, considering the influence of Gender as the protected attribute. As shown in Table \ref{tab:lfw_results}, our findings on our proposed approach on the LFW dataset \cite{LFWTech} are consistent with those on the CelebA dataset \cite{liu2015faceattributes}, indicating an increase in overall accuracy and improvements across most fairness metrics.

\paragraph{\textbf{Comparing the proposed approach across different backbones}}Table \ref{tab:backbone_results} presents a quantitative comparison of our approach using different backbones (ResNet-18 \cite{he2016deep}, ResNet-50\cite{he2016deep}, DenseNet-121\cite{huang2017densely}, and DenseNet-169\cite{huang2017densely}) on the CelebA dataset \cite{liu2015faceattributes}. Similar to our evaluation on the CelebA and LFW datasets, we use the FGSM attack with $\epsilon=\{0,0.01,0.001\}$ and to construct the ASACs in each minibatch and $\alpha=0.5$ in the loss function (Equation \ref{eq:adv_loss}). Our method obtains significant improvements in accuracy while also improving the fairness of the model, showing that our method generalizes across architectures and scale of the model.

\subsection{Qualitative Results}
\label{sub:qual_results}

\subsubsection{Robustness evaluation of adversarial training}
\label{subsub:robust_eval}


To assess the impact of adversarial training using ASACs, we examine how adversarial images designed to attack a gender classifier affect the decision of the smile classifier, pre and post-training with our proposed approach. Given an image, we add varying magnitudes of adversarial noise and provide the corrupted images to the smile classifier. We observe that before using our proposed method, the adversarial noise intended to mislead the gender classifier also misled the smile classifier, indicating a possible correlation between gender and smile attributes in these examples. Such an outcome suggests that manipulations the target gender can influence predictions on e.g., smile prediction. We perform an identical analysis on the models after our proposed fine-tuning approach. We observe that after fine-tuning, the target classifier improves its discriminative ability while the protected attribute classifier used to construct ASACs continues to be deceived. This indicates that the fine-tuning procedure helps to decouple spurious correlations between target and protected attribute. This enables the target classifier to make more accurate predictions independent from the protected attribute. Qualitative examples of these results have been moved to the Supplementary section (Figure \ref{fig:rob_postive} ) due to space constraints. 

\paragraph{\textbf{Integrated Gradients}}
\label{subsub:integrated_grads}

\begin{figure}[tb]
  \centering
   \includegraphics[width=\linewidth]{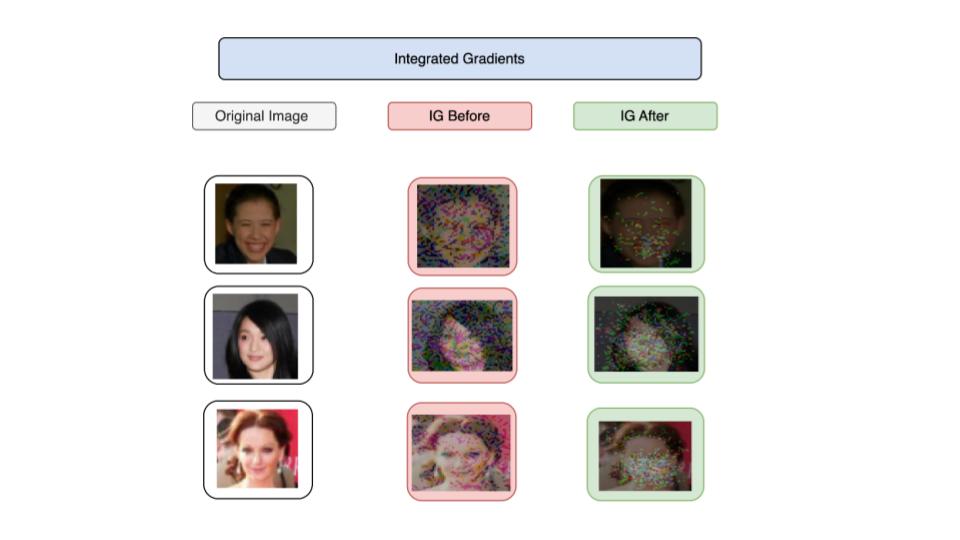}
   \caption{We look at the Integrated Gradients and for samples pre (IG before) and post (IG after) training.}
   \label{fig:ig_tsne}
\end{figure}

To assess the impact of ASAC-driven adversarial and curriculum-based fine-tuning, we observe the effect our approach has on individual samples in the test set using Integrated Gradients (IG) \cite{sundararajan2017axiomatic}. Integrated Gradients is a widely recognized interpretability technique that highlights pixels influencing decision-making process of the model. In Figure \ref{fig:ig_tsne}, we juxtapose the integrated gradients of decisions made by the classifier in both pre and post ASAC training settings. For more examples, see Section \ref{sup:add_IG_resutls} in the supplementary material. After applying our proposed approach, the integrated gradients converge towards the mouth area in the images. As humans, we can agree that the mouth is the largest indicator of whether a person is smiling or not. Our observations suggest that ASAC-driven training compel the model to focus on regions of the image more closely correlated with a target attribute, such as smile in this case. Consequently, these findings indicate that the model fine-tuned using ASACs may give more importance to regions closer to the mouth compared to the original classifier (pre fine-tuning).

\subsection{Ablation Study}
\label{sub:ablations}

In our ablation study, we use a ResNet-18 backbone with the target attribute as smile and the protected attribute as gender. We utilize FGSM with $\epsilon=\{0,0.01,0.001\}$ to generate the ASACs for each minibatch on the CelebA dataset and $\alpha=0.5$ for the loss function (Equation \ref{eq:adv_loss}).

\begin{table}[t]
\caption{A comparison of different curriculum learning strategies.}
\label{tab:cl_order}
\vspace*{-0.15in}
\begin{center}
\begin{small}
\begin{tabular}{lrrrr}
\toprule
Method & \hspace{3mm} DEO & \hspace{3mm} DEOp & \hspace{3mm} DDP & \hspace{3mm} ACC \\
\midrule
W/O Curriculum Learning & 0.055	&0.044	&\textbf{0.14}	&91.23\\
Curriculum Learning in Ascending order&\textbf{0.050}	&\textbf{0.043}	& 0.15	&91.79\\
Curriculum Learning in Descending order  &0.058&	0.048&	0.17	&\textbf{92.08}
 \\
\bottomrule
\end{tabular}
\end{small}
\end{center}
\vskip -0.1in
\end{table}
\begin{table}[t]
\caption{Ablation on the effect of different values of noise magnitudes ($\epsilon$) on the performance and fairness metrics of the target classifier.}
\label{tab:noise_mags_alpha_ablation}
\vspace*{-0.15in}
\begin{center}
\begin{small}
\begin{tabular}{lcrrrr}
\toprule
Parameter & \hspace{3mm} Setting & \hspace{3mm} DEO & \hspace{3mm} DEOp & \hspace{3mm} DDP& \hspace{3mm} ACC \\
\midrule
$\epsilon$ & \{0.001, 0.01\}&\textbf{0.050} &\textbf{0.043}	&0.15	&\textbf{91.91}\\
 & \{0.001,0.03\}&0.055	&0.043	&\textbf{0.14}	&91.42\\
 & \{0.001,0.05\}&0.057	&0.047 &0.14	&91.38\\
\midrule
\end{tabular}
\end{small}
\end{center}
\vskip -0.1in
\end{table}

\textbf{Evaluating the impact of Curriculum Learning}
In this study, we evaluate the models performance over three training strategies: curriculum learning where the order of ASACs in a minibatch is random, where the difficulty (confidence of the model in misclassification) gradually increases, and starting with the most difficult examples i.e., the difficulty decreases. Results (Table \ref{tab:cl_order}) show that while beginning with challenging examples boosts accuracy, saving them for later enhances fairness metrics. This highlights the importance of curriculum design in balancing accuracy and fairness.



\paragraph{\textbf{Finding the most optimal noise magnitudes}}
We examine the effect varying noise magnitudes on the performance of our smile classifier (Table \ref{tab:noise_mags_alpha_ablation}). This study provides insights into selecting noise magnitudes to enhance model robustness and effectiveness. We find that noise magnitudes of $\epsilon = 0.03$ and $\epsilon = 0.05$ yield optimal results. Further experiments are detailed in the supplementary material (Section \ref{sup:noise_mags_alpha_ablation2}).

\section{Discussion and Conclusion}
\label{sec:discussion}


We proposed a method for utilizing attribute-specific adversarial examples as counterfactuals to enhance fairness in computer vision models. Our method fine-tunes models with these counterfactuals using a novel curriculum learning training regime. Our experiments on CelebA and LFW datasets demonstrate that adversarial fine-tuning can significantly improve fairness metrics without sacrificing accuracy. Qualitative results indicate the model's potential ability to disentangle predictions  from protected attributes. Ablation studies justify different design choices, notably revealing that starting with challenging examples may lead to a more accurate model, while presenting easier examples earlier results in a fairer but less accurate model. This underscores the impact of curriculum learning on the problem. For our future work, we plan to investigate the impact of various adversarial attacks on the model. Additionally, we aim to explore how adversarial images can be utilized to interpret the relationship between model decisions and the underlying classifiers.

\paragraph {\textbf{Ethical Considerations.}}
Our work targets the ethical concerns surrounding biases in computer vision models. Further, our approach can be used to mitigate biases by fine-tuning existing biased models and can also deployed in a low-data setting. The main limitations of our work are that (1) it requires the backbone of our model and cannot be applied on a black-box model, and (2) it relies on knowledge of the protected features, which is sometimes inaccessible. (3)Finally, we only assess gender as a sensitive attribute. The rationale behind this experimental choice is detailed in the Supplementary material \ref{sup:bias_exp}

\vspace{-0.2cm}
\section{Acknowledgements}
This work was supported by the Simons Foundation Collaboration on the Theory of Algorithmic Fairness, the Sloan Foundation Grant 2020-13941, and the Simons Foundation investigators award 689988. The authors would like to thank these organizations for their generous support, which has been instrumental in advancing our research.

\bibliographystyle{abbrv}
\bibliography{ref}

\newpage
\appendix

\section{Supplementary Material}

\subsection{Fairness Definitions}
\label{sup:Deifintion}
To evaluate the fairness of our model, we employ three key metrics: the Difference in Demographic Parity (DDP), the Difference in Equalized Odds (DEO), and the Difference in Equalized Opportunity (DEOp). Alongside these, we also measure the model's accuracy (ACC).  

The DEO focuses on the difference in the rates of false negatives and false positives between genders. A substantial DEO indicates a bias in which one group is either less likely to be incorrectly dismissed or more likely to be inaccurately favored. 

\begin{definition}[Equalized Odds]
A predictor $\hat{Y}$ satisfies equalized odds with respect to protected attribute $A \in \{a,a'\}$ and outcome $Y$, if $\hat{Y}$ and $A$ are independent conditional on $Y$.
\begin{equation}
P(\hat{Y}=1|Y=y,A=a)=P(\hat{Y}=1|Y=y,A=a')  \quad \text{where } y \in \{0,1\}.
\end{equation}
To estimate how far a predictor $\hat Y$ from having equalized odds, we  estimate the Difference in Equalized Odds (DEO), i.e., we estimate
\[
\frac{\sum_{y \in \{0,1\}}|P(\hat{Y}=1|Y=y,A=a)-P(\hat{Y}=1|Y=y,A=a')|}{2}.
\]
\end{definition}

DDP quantifies the absolute gap in approval rates (e.g., smile prediction) across different protected attributes.  A large DDP value suggests a tendency for individuals in a specific group to receive more favorable outcomes than their counterparts in the other group. 
\begin{definition}[Demographic Parity]
A predictor $\hat{Y}$ satisfies demographic parity with respect to protected attribute $A \in \{a, a'\}$ and outcome $Y$, if the following condition holds:
\begin{equation}
P(\hat{Y} = 1 | A = a) = P(\hat{Y} = 1 | A = a').
\end{equation}
To estimate how far  a predictor $\hat Y$ from having demographic parity, we  estimate the Difference in Demographic Parity (DDP), i.e., we estimate
\[
|P(\hat{Y}=1|A=a)-P(\hat{Y}=1|A=a')|.
\]
\end{definition}

The Difference in Equalized Opportunity (DEOp) measures the disparity in True Positive Rates between different demographic groups in a model, indicating bias in favorable outcomes. A DEOp of zero signifies equal accuracy across groups, representing a fairness ideal in model performance. The difference in Equalized Opportunity is given as follows. 
\begin{definition}[Equalized Opportunity]
A predictor $\hat{Y}$ satisfies equalized opportunity with respect to protected attribute $A \in \{a, a'\}$ and outcome $Y$, if the following condition holds for a specific outcome value $y$:
\[
P(\hat{Y}=1 | Y=1, A=a) = P(\hat{Y}=1 | Y=1, A=a').
\]
To measure how far a predictor $\hat Y$ from having demographic parity, we  estimate the Difference in Equalized Opportunity (DEOp), i.e., we estimate
\[
|P(\hat{Y}=1 | Y=1, A=a) - P(\hat{Y}=1 | Y=1, A=a')|.
\]
\end{definition}
Ideally, we want these values to be zero, indicating no disparity. 

\section{Additional Ablation Studies}
\begin{table}[t]
\caption{Ablation on the effect of different values of $\epsilon$ and $\alpha$ on the performance and fairness metrics of the target classifier utilizing FGSM for ASAC generation.}
\label{sup:noise_mags_alpha_ablation2}
\vspace*{-0.15in}
\begin{center}
\begin{small}
\begin{tabular}{lcrrrr}
\toprule
Parameter & \hspace{3mm} Setting & \hspace{3mm} DEO & \hspace{3mm} DEOq & \hspace{3mm} DDP& \hspace{3mm} ACC \\
\midrule
$\epsilon$ & \{0.001,0.03,0.05\}&0.065	&0.047	&0.16	&91.00\\ 
 & \{0.001\}&0.054	&0.043	&0.14	&91.62\\
 & \{0.05\}&0.055	&0.044	&0.14	&91.71\\
\midrule
\midrule
$\alpha$ & 0.3 &0.06	&0.04	&0.16& 92.04  \\
 & 0.5  &0.058 &0.050 &	0.16&	91.84 \\
 & 0.7 &0.057&0.047	&0.14&91.90 \\
\bottomrule
\end{tabular}
\end{small}
\end{center}
\vskip -0.1in
\end{table}
\paragraph{\textbf{Choosing an Optimal $\alpha$}}
\begin{figure}
  \centering  
   \includegraphics[width=\linewidth]{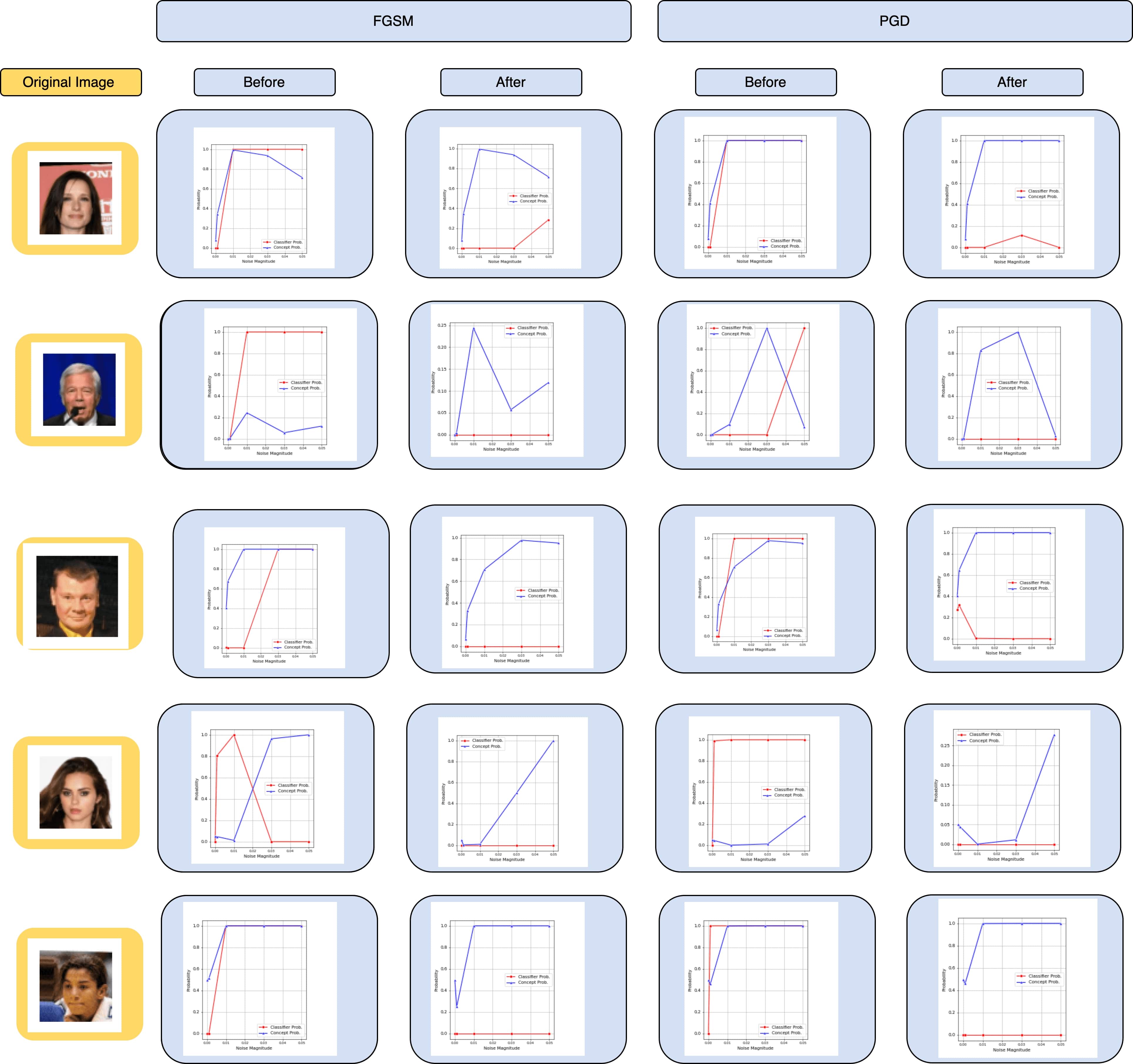}
   \caption{Qualitative results showing that our trained model becomes robust to ASACs after training.}
\label{fig:rob_postive}
\end{figure}
\begin{figure}
  \centering
   \includegraphics[width=\linewidth]{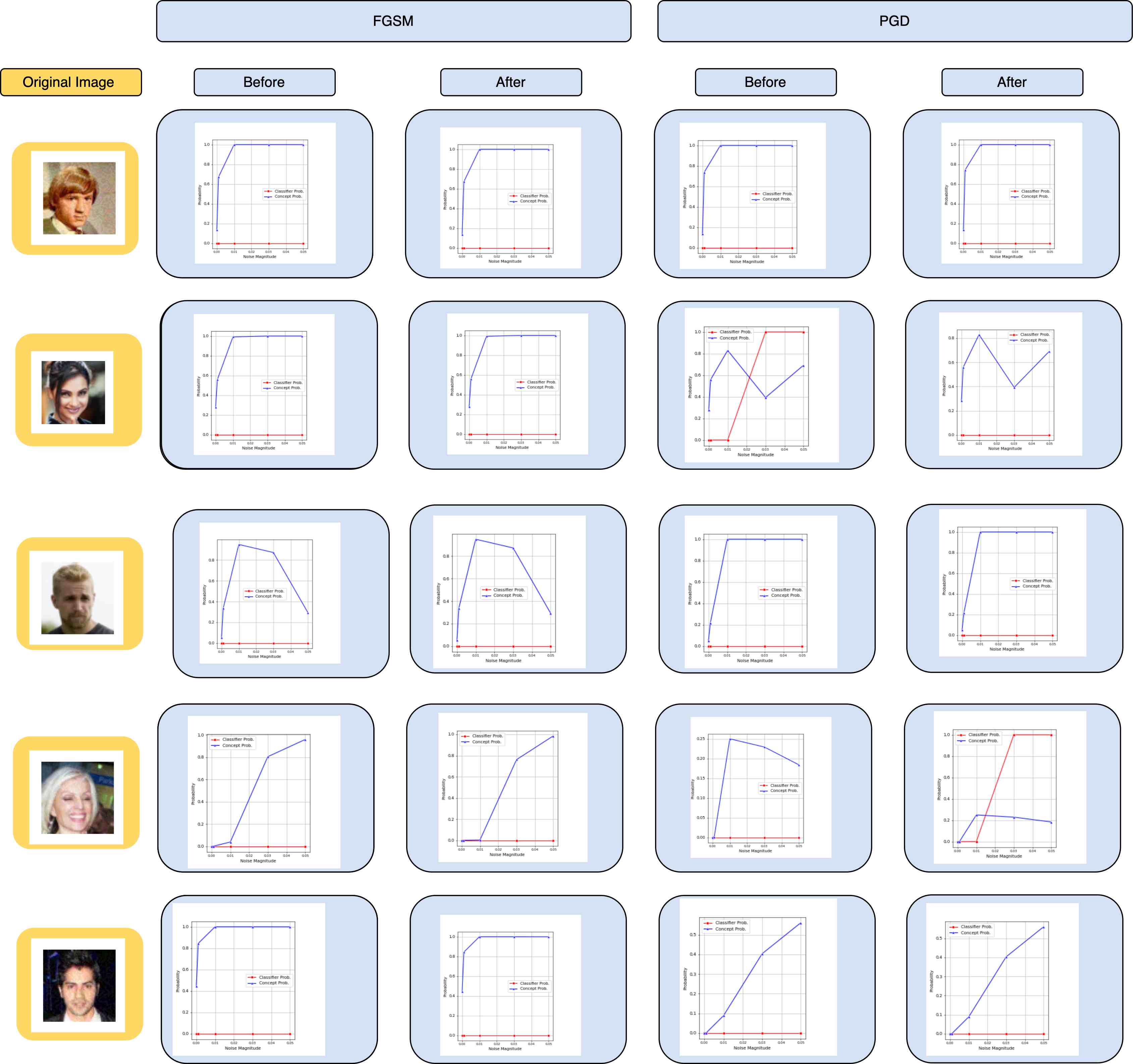}
   \caption{Examples where the adversarial noise is unable to flip the smile classifier. As shown in the figure adding adversarial noise does not change the decision of the smile classifier (red curve). Post training, the decision remains unchanged as well.}
\label{fig:rob_negative}
\end{figure}
\begin{figure}
  \centering
   \includegraphics[width=\linewidth]{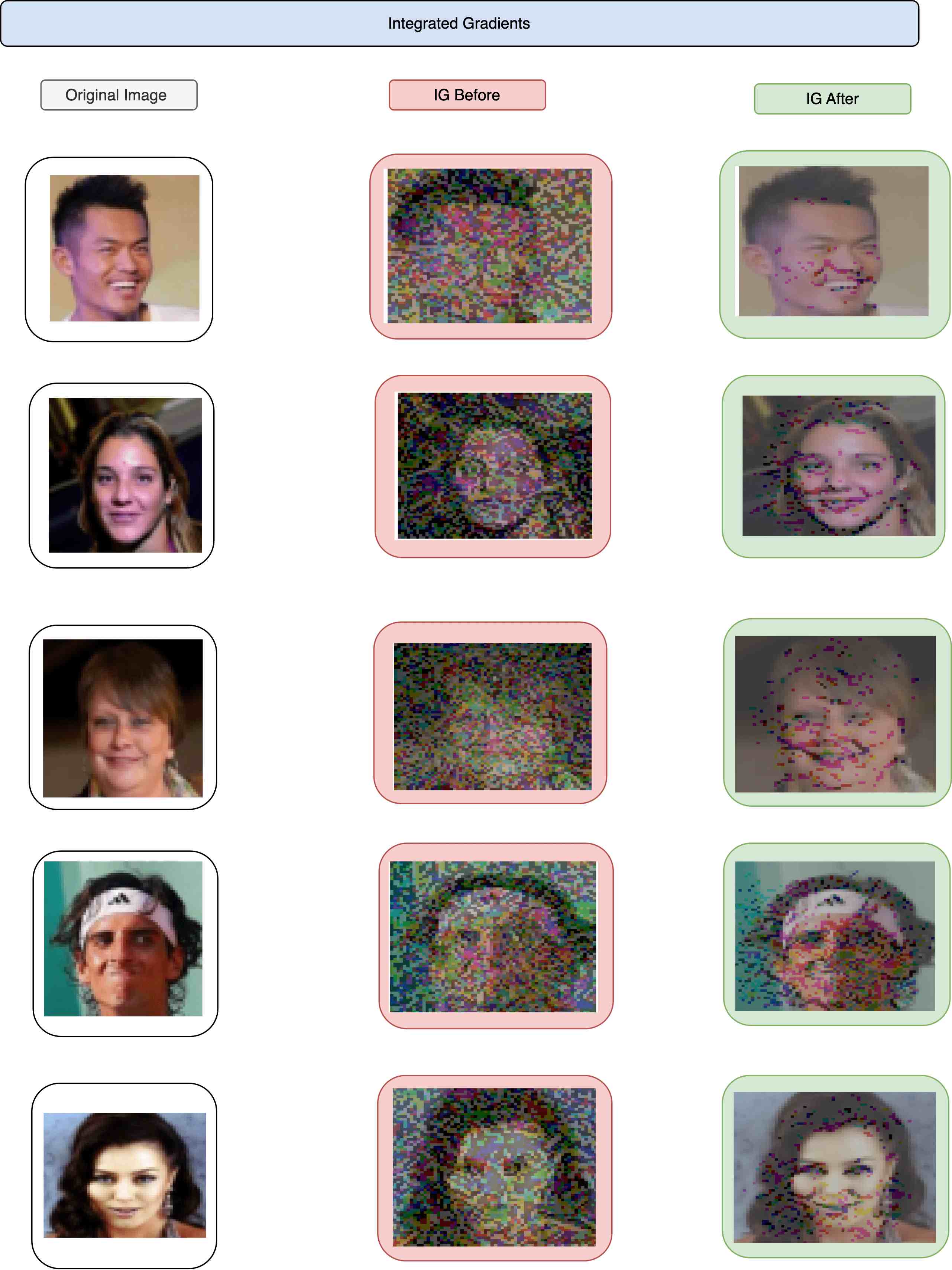}
   \caption{More examples showcasing the change in integrated gradients for the smile classifier before and after applying our method. }
\label{fig:add_IG_results}
\end{figure}
\label{sup:add_IG_resutls}
In our third ablation study (Table \ref{sup:noise_mags_alpha_ablation2}), we examine the impact of different $\alpha$ values ($0.3$, $0.5$, and $0.7$) on model performance. Lower $\alpha$ ($0.3$) leads to higher accuracy, whereas $\alpha=0.5$ prioritizes fairness metrics. We select $\alpha=0.5$ for final model training, emphasizing the importance of balancing accuracy and fairness. Though a lower value of $\alpha$ boosts accuracy, $\alpha=0.5$ achieves favorable fairness outcomes, ensuring a more equitable model. These experiments are also conducted on a base ResNEt18 smile classifier trained on CelebA data.
\paragraph{\textbf{Finding optimal noise magnitudes}}
This section shows some additional results for different noise magnitude combinations. These results have been shown in Table (\ref{sup:noise_mags_alpha_ablation2}
\subsection{Additional Robustness Results}
This section shows some additional positive (Figure \ref{fig:rob_postive}) and negative results (Figure \ref{fig:rob_negative}) similar to the qualitative results. shown in Section \ref{sub:qual_results}. It should be noted that however, some samples are not impacted by adding adversarial noise and as a result, the values of the smile classifier do not change. These examples have been shown in Figure \ref{fig:rob_negative}.

\subsection{Additional IG Results}
In Figure \ref{fig:add_IG_results}, we show more examples of how Integrated Gradients (IG) change on sample pictures after we use our new method. Integrated Gradients help us see which parts of the picture affect the model's decisions the most. We did our tests the same way we explained in Section \ref{subsub:integrated_grads}. Examples presented in Figure \ref{fig:add_IG_results} show a similar trend to Figure \ref{fig:ig_tsne} and the pixels that are used in decision-making converge from the entire towards the mouth region after applying our approach. 


\subsection{Comparison along other dimensions of bias}
\label{sup:bias_exp}
In this paper, we focus on gender as the primary dimension of bias. The primary reason for this choice is the availability of data and the clear presence of spurious correlations. For instance, in the CelebA dataset, both gender and smile attributes have nearly uniform distributions and show a positive pearson correlation of $0.23$ . In contrast, other attributes like age in CelebA exhibit a highly skewed distribution (22$\%$ young and 78$\%$ old) and weak correlations with other uniformly distributed attributes (0.12 with smile, 0.11 with big nose, and 0.16 with wavy hair). Additionally, the CelebA dataset does not include any attributes related to race similarly the LFW data set does not have one label for race but multiple such labels such Indian, Black, White etc which makes it harder for us to draw a comparison. Consequently, our analysis is limited to gender bias.

\subsection{Training setup}
\label{sup:train_setup}
For our experiments, we used a training cluster utilizing a single Nvidia RTX A6000 GPU with 48 GB of GDDR6 memory, 384GB of RAM and 20 CPU cores. Each training/fine-tuning run took between 60 to 120 minutes to run. In addition, for experiments involving ResNet-18, we used an Apple M1 Pro chip with an 8-core CPU, 14-core GPU, 16-core Neural Engine and 16GB of RAM. For experiments of the same scale, we also used a Nvidia A100 with 40GB of memory and 25GB of RAM. Unless otherwise mentioned, we train and fine-tune models using a ResNet-18 backbone, with $\alpha=0.5$ and a noise magnitude setting of $\{0.01,0.001\}$. We train the base classifiers for 50 epochs and fine-tune using our curriculum learning approach for 10 epochs with both stages employing a learning rate of $1 \times 10^{-3}$. Across all experiments we use the Adam \cite{kingma2014adam} optimizer with $\beta_1=0.9$, $\beta_2=0.999$ and $\epsilon=1 \times 10^{-8}$ and perform gradient clipping with a value of $1.0$.


\newpage

\end{document}